\documentclass[graybox]{svmult}

\usepackage{mathptmx}       
\usepackage{helvet}         
\usepackage{courier}        
\usepackage{type1cm}        
\usepackage{makeidx}         
\usepackage{multicol}        
\usepackage[bottom]{footmisc}

\usepackage[utf8]{inputenc} 
\usepackage[T1]{fontenc}    
\usepackage{hyperref}       
\usepackage{url}            
\usepackage{booktabs}       
\usepackage{amsfonts}       
\usepackage{nicefrac}       
\usepackage{microtype}      
\usepackage[normalem]{ulem}
\usepackage{cite}

\usepackage{bm}
\usepackage[cmex10]{amsmath}
\usepackage[vlined,ruled]{algorithm2e}
\usepackage{algpseudocode}
\usepackage[mathscr]{euscript}

\usepackage[pdftex]{graphicx}
\graphicspath{{./figures/}}
\DeclareGraphicsExtensions{.pdf,.jpeg,.png}
\usepackage[caption=false]{subfig}

\usepackage{fixme}
\fxsetup{%
    status=draft,
    theme=color
}

\newcommand*{\argmax}[1]{\underset{#1}{\mathrm{argmax}~}}
\newcommand*{\argmin}[1]{\underset{#1}{\mathrm{argmin}~}}
\newcommand*{\transpose}{\mathrm{T}}

\newcommand*{\dataset}{\mathscr{D}} 
\newcommand*{\slocation}{x} 
\newcommand*{\location}{\mathbf{\slocation}} 
\newcommand*{\locSet}{\mathbf{X}} 

\newcommand*{\locMean}{\mathbf{u}}
\newcommand*{\locNoise}{\epsilon}

\newcommand*{\observation}{z} 
\newcommand*{\observations}{\mathbf{z}}
\newcommand*{\fValue}{y} 
\newcommand*{\fValues}{\mathbf{y}}
\newcommand*{\nObs}{n}

\newcommand*{\gp}{{GP}}
\newcommand*{\normal}{\mathscr{N}}
\newcommand*{\af}{h}
\newcommand*{\expectation}{\mathbb{E}}

\newcommand*{\R}{\mathbb{R}}

\newcommand*{\Sspace}{\mathscr{S}}
\newcommand*{\pMeasure}{\mathbb{P}}

\title*{Bayesian Optimisation for Safe Navigation under Localisation Uncertainty}
\author{Rafael Oliveira, Lionel Ott, Vitor Guizilini and Fabio Ramos}
\institute{The University of Sydney, Australia, \email{rdos6788@uni.sydney.edu.au}
}

\begin{document}

\maketitle              

\abstract{
In outdoor environments, mobile robots are required to navigate through terrain with varying characteristics, some of which might significantly affect the integrity of the platform. Ideally, the robot should be able to identify areas that are safe for navigation based on its own percepts about the environment while avoiding damage to itself. Bayesian optimisation (BO) has been successfully applied to the task of learning a model of terrain traversability while guiding the robot through more traversable areas. An issue, however, is that localisation uncertainty can end up guiding the robot to unsafe areas and distort the model being learnt. In this paper, we address this problem and present a novel method that allows BO to consider localisation uncertainty by applying a Gaussian process model for uncertain inputs as a prior. We evaluate the proposed method in simulation and in experiments with a real robot navigating over rough terrain and compare it against standard BO methods.}



\section{Introduction}
\label{sec:int}

Mobile robots have been successfully applied to many field applications, such as mining \cite{Maekawa2010}, planetary exploration \cite{TobiasLangChristianPlagemann2007}, agriculture \cite{Underwood2017}, and environmental monitoring \cite{Marchant2012}, to name a few. In all these applications, robots face environments with physical characteristics that are {\em a priori} unknown and can heavily affect performance. In the case of ground robots, terrain roughness can affect the ability of a robot to navigate and even cause damage to its on-board hardware due to excessive vibration\cite{Souza2014}. To aid in these problems, methods to enable the robot to automatically learn terrain properties from its sensory data have been presented in the literature \cite{TobiasLangChristianPlagemann2007,Souza2014,Oliveira2016}. However, such methods usually assume that localisation is accurate enough, without dealing with its inherent uncertainty.


Uncertainty in localisation can mislead learning algorithms with noise-corrupted estimates of the location where measurements are taken. In addition, localisation accuracy also affects navigation, since local path execution heavily depends on knowing where the robot is with respect to a given reference path. As a result, localisation inaccuracy can lead a robot into areas that are unsafe for navigation.


In this paper, we propose a method that allows a robot to actively search for, and learn a model of, traversable areas while keeping itself safe by considering both model and localisation uncertainty. In particular, we propose:
\begin{itemize}
\item a framework to account for input uncertainty in Bayesian optimisation \cite{Brochu2010} for mission planning; and
\item an adaptation of the DUCB \cite{Marchant2012} acquisition function to the context of exploration under uncertain localisation.
\end{itemize}
In our method, we apply an extension of the conventional Gaussian process (GP) \cite{Rasmussen2006} regression models to contexts involving input uncertainty \cite{Girard2004} as priors for the BO framework. Such GP models allow BO to take into account noise in both the execution of a query and in the observation's location estimate. We apply the proposed method to the task of learning a model of terrain roughness from experienced vibration data, as in \cite{Souza2014}.

The remainder of this paper is organised as follows.
In the next section, we review relevant prior work in the areas of terrain modelling and Bayesian optimisation. Section \ref{sec:pre} revises the general GP-based BO framework, which does not consider uncertain inputs. In Section \ref{sec:met} we present our method for Bayesian optimisation under localisation uncertainty. Then, in Section \ref{sec:exp}, we present experimental results in simulation and on a physical robot to evaluate our approach. Finally, in Section \ref{sec:con}, we conclude and propose some directions for future work.

\section{Related Work}
\label{sec:rel}

Traversability metrics estimate how hard it is for a vehicle to traverse a certain terrain. In robotics, terrain traversability is usually estimated from either LIDAR range measurements \cite{Nordin2012} or stereo vision information \cite{Ho2013}. 
Among the different kinds of metrics, terrain roughness can play an important role in planning how a robot should navigate on the terrain \cite{Nordin2012}. Methods to learn terrain roughness using image data usually rely on learning image classification models \cite{Komma2009}, which generally do not provide a measure of how uncertain they are about their estimates and are computationally intensive. In \cite{Martin2013}, however, the authors propose using vehicle experience data to learn a Gaussian process (GP) regression model \cite{Rasmussen2006} to predict terrain traversability with uncertainty estimates. Though restrictive when compared to exteroceptive sensing information \cite{Ho2013}, proprioceptive sensing can still be a very viable option for traversability estimation with robots that are not equipped with stereo vision or 3D LIDAR sensors, or do not possess the computational resources to process this kind of information on-board online.

Souza et al. \cite{Souza2014} presented a method to learn a GP model for terrain roughness from vehicle experience. The authors applied Bayesian optimisation (BO) \cite{Brochu2010} in an active perception approach to reduce experienced vibration during navigation while learning the model from IMU measurements online. In the BO framework, the terrain roughness model is learnt online as the algorithm drives the robot around selecting locations to visit balancing a trade-off between exploration and exploitation \cite{Brochu2010,Souza2014}. Nevertheless, BO usually considers deterministic query locations within its search space, as BO is typically used in problems where a fixed number of parameters have to be optimised \cite{Snoek2012,Wilson2014}.

Recent work \cite{Nogueira2016} presented a method to apply BO to problems where the execution of a query is uncertain, such as robotic grasping. The authors propose querying BO's surrogate model using a Gaussian distribution by applying the unscented transform \cite{Wan2000}. In this, even with uncertainty in the execution of the query, the algorithm chooses to sample the objective function at locations where interesting values are more likely to be observed, instead of trying to reach a narrow peak. Despite that, \cite{Nogueira2016} still applies a deterministic-inputs GP model as a prior for BO. In navigation problems, the robot is usually able to obtain a probability distribution estimating its location from a localisation system. Making use of a GP model that takes into account such distributions as inputs should then allow BO to learn a better model of the true underlying objective function.

To deal with uncertain location estimates in observations, besides the usual noisy outputs, BO needs a statistical model that considers partial observations of the inputs when sampling the function being modelled, usually called the \textit{objective} function. In BO the most common approach to modelling the objective function is using Gaussian process (GP) regression \cite{Rasmussen2006}. In the case of uncertain input observations, the work in \cite{Girard2004} proposed methods to propagate input uncertainty through a GP model, developing analytical approximations to compute covariance function values between inputs represented as Gaussian distributions. Another approach is considered in \cite{Mchutchon2011}, where the input location estimates are assumed to be corrupted by i.i.d. Gaussian noise, similar to the output observations in standard GP regression. The authors present a method to propagate the input uncertainty to the output of the model using a first-order Taylor approximation. Finally, a more general framework is presented in \cite{Damianou2016}, where the true unobserved inputs of the GP model are considered as latent variables and a variational inference framework is applied to compute the posterior of this GP model under a set of assumptions about the input distributions. None of these methods, however, have been applied to the BO context, where the GP model is built in an online fashion as the algorithm proceeds.

\section{Preliminaries}
\label{sec:pre}
In this section, we review the concepts that form the basis of the method we propose. We start with a general introduction to Gaussian process (GP) regression, which is used to model terrain properties. Then we follow with a short review of Bayesian optimisation (BO) using GP priors, which will be our planning technique.


\subsection{Gaussian process regression}
Gaussian process regression \cite{Rasmussen2006} is a Bayesian non-parametric framework that places a Gaussian distribution as a prior over the space of functions $f$, mapping inputs $\location \in \mathbb{R}^d$ to outputs $\fValue \in \mathbb{R}$. In a similar way to a conventional Gaussian distribution, a GP model needs to specify the mean and the covariance for any given pair of values in its vector space, which for a GP is a reproducing kernel Hilbert space (RKHS) \cite{Rasmussen2006,Aronszajn1950}. Considering a mean function $\mu_0: \mathbb{R}^d \to \mathbb{R}$ and a positive-definite covariance function $k:\mathbb{R}^d\times\mathbb{R}^d \to \mathbb{R}$, a GP prior models the distribution of the function values $\fValues = [\fValue_1,\dots,\fValue_\nObs]^\transpose$ at a given set of input locations $\locSet = \{\location_i\}_{i=1}^\nObs$ as:
\begin{equation}
\fValues|\locSet \sim \normal(\mu_0(\locSet),k(\locSet,\locSet)) ~,
\label{eq:gp_prior}
\end{equation}
where $\mu_0(\locSet)=[\mu_0(\location_1),\dots,\mu_0(\location_\nObs)]^\transpose$ and $[k(\locSet,\locSet)]_{ij} = k(\location_i,\location_j)$. In many practical applications the prior mean $\mu_0$ is set to zero or a constant that can be learnt from a dataset. For the covariance function there are a few popular choices, one of them is the squared exponential kernel:
\begin{equation}
\label{eq:k_se}
k(\location,\location') = \sigma_f^2 \exp(-\frac{1}{2}(\location-\location')^\transpose \mathbf{W}^{-1} (\location-\location')) ~,
\end{equation}
where $\sigma^2_f$ is a signal variance parameter and $\mathbf{W}$ is a $d\times d$ diagonal matrix with $[\mathbf{W}]_{ii} = \lambda_i^2$, and each $\lambda_i$ is a length-scale parameter indicating how much the function can vary along the $i$-th dimension. 

Given a set of observations $\dataset = \{\location_i,\observation_i\}$, containing both noise-corrupted outputs $\observation_i = f(\location_i) + \epsilon_i$ with $\epsilon_i \sim \normal(0,\sigma_n^2)$ and deterministic input locations $\location_i$, the values for $\fValue_* = f(\location_*)$ at a given query location $\location_*$ are then distributed according to a Gaussian posterior:
\begin{equation}
\fValue_*|\observations,\locSet,\location \sim \normal(\mu(\location_*),\sigma^2(\location_*)) ~,
\label{eq:gp_posterior}
\end{equation}
where:
\begin{eqnarray}
\mu(\location_*) &=& \mu_0(\location_*) + {k}(\location_*,\locSet)K_\locSet^{-1}(\observations-\mu_0(\locSet)) \label{eq:mean}\\
\sigma^2(\location_*) &=& k(\location_*,\location_*) - {k}(\location_*,\locSet)K_\locSet^{-1}{k}(\locSet,\location_*) ~, \label{eq:var}
\end{eqnarray}
using ${K_\locSet} = k(\locSet,\locSet) + \sigma_n^2I$. Therefore, GP models can be used as priors to model functions that are not directly observable during an optimisation process. Partial observations $z_t$ can be collected from the function being optimised to incrementally update the GP model. If the GP prior is appropriately specified, as more observations are collected the variance in the posterior decreases, leading the mean to converge on top of the true $f$.


\subsection{Bayesian optimisation}
Consider the problem of searching for the global optimum of a function $f:\mathbb{R}^d \to \mathbb{R}$ in a certain compact region $\Sspace \subset \mathbb{R}^d$, i.e.:
\begin{equation}
\location^* = \argmin{\location \in \Sspace} f(\location) ~.
\label{eq:objective}
\end{equation}
Assume that $f$ is unknown to us and we have only access to noisy observations of its output $\observation = f(\location) + \epsilon$ with $\epsilon \sim \normal(0,\sigma_n^2)$ and that we can only sample the function $N$ times.

Bayesian optimisation \cite{Brochu2010} assumes that $f$ is a random variable itself and applies a probability distribution over it using a statistical model. Using a Gaussian process (GP) model, BO makes some prior assumptions about $f$ encoded by the mean $\mu_0(\location)$ and covariance function $k$.

Rather than directly searching over $f$, BO uses an \emph{acquisition function} $\af(\location)$ as a guide to sequentially select input locations at which to observe $f$. The acquisition function uses the information provided by the GP prior and the observations of $f$ to estimate an utility value for sampling $f$ at a given $\location$. So at each iteration, BO queries the objective $f$ at the location of highest utility according to the acquisition function,  $\location_t$, such that:
\begin{equation}
\location_t = \argmax{\location \in \Sspace} h(\location) ~.
\end{equation}
After observing $f$ at $\location_t$, BO updates the GP's observations dataset with the new observation $\{\location_t,\observation_t\}$, which improves its belief about $f$. After the GP update the algorithm proceeds to the next iteration, choosing an $\location_{t+1}$. The BO loop runs until a stopping criteria is satisfied, which is usually defined by a maximum budget of iterations. In the end of this process, BO obtains a model that approximates the objective function $f$ and an estimate of its optimum's location $\location^*$.

One example of acquisition function that can be applied to problems involving robotics navigation is the distance-based upper confidence bound (DUCB) \cite{Marchant2012}:
\begin{equation}
\af_{\text{DUCB}}(\location|\dataset_{t-1}) = \mu_{t-1}(\location) + \kappa \sigma_{t-1}(\location) - \gamma d(\location_{t-1},\location) ~,
\end{equation}
where $d(\location_{t-1},\location)$ corresponds to the distance between the last sampled location and the candidate $\location$, $\mu_{t-1}$ and $\sigma_{t-1}$ are given by the GP posterior with the observations in $\dataset_{t-1}$, and $\kappa >  0$, $\gamma > 0$ are parameters to be set. In this sense, $\kappa$ controls the exploration-exploitation trade off, with higher values favouring areas of high uncertainty, while $\gamma$ penalises large jumps, allowing shorter paths between observations. In addition, notice that for minimisation objectives, as in Equation \ref{eq:objective}, DUCB can be applied by simply flipping the sign of the GP posterior mean to negative instead.

When the hyper-parameters of the GP model are unknown a priori, a few methods can also be applied to adapt them online as BO runs. One of the simplest, but usually effective, of them is to optimise the log-marginal likelihood of the GP \cite{Rasmussen2006} after each observation or after collecting a batch of observations. This optimisation can be done by performing a few steps of gradient descent on the hyper-parameters after the GP is updated. 

\section{Bayesian optimisation under localisation uncertainty}
\label{sec:met}
In this section we present our framework for Bayesian optimisation under uncertain inputs. By inputs, we refer to points in the objective function $f$ (Equation \ref{eq:objective}) input space, i.e. any $\location \in \R^d$. Input noise can affect both the execution of a query to observe the objective function and the location estimates of where these observations are taken, as we further explain in Section \ref{sec:uibo}. In this sense, we first need to extend BO's surrogate model to incorporate uncertainty in the inputs. We propose to use a Gaussian process regression model with uncertain inputs \cite{Girard2004} as a prior, which we then incorporate into BO to obtain a framework for optimisation under uncertain inputs. We start by explaining this GP model before going into details about the proposed BO approach.

\begin{figure}[t]
\centering
\includegraphics[width=0.8\textwidth]{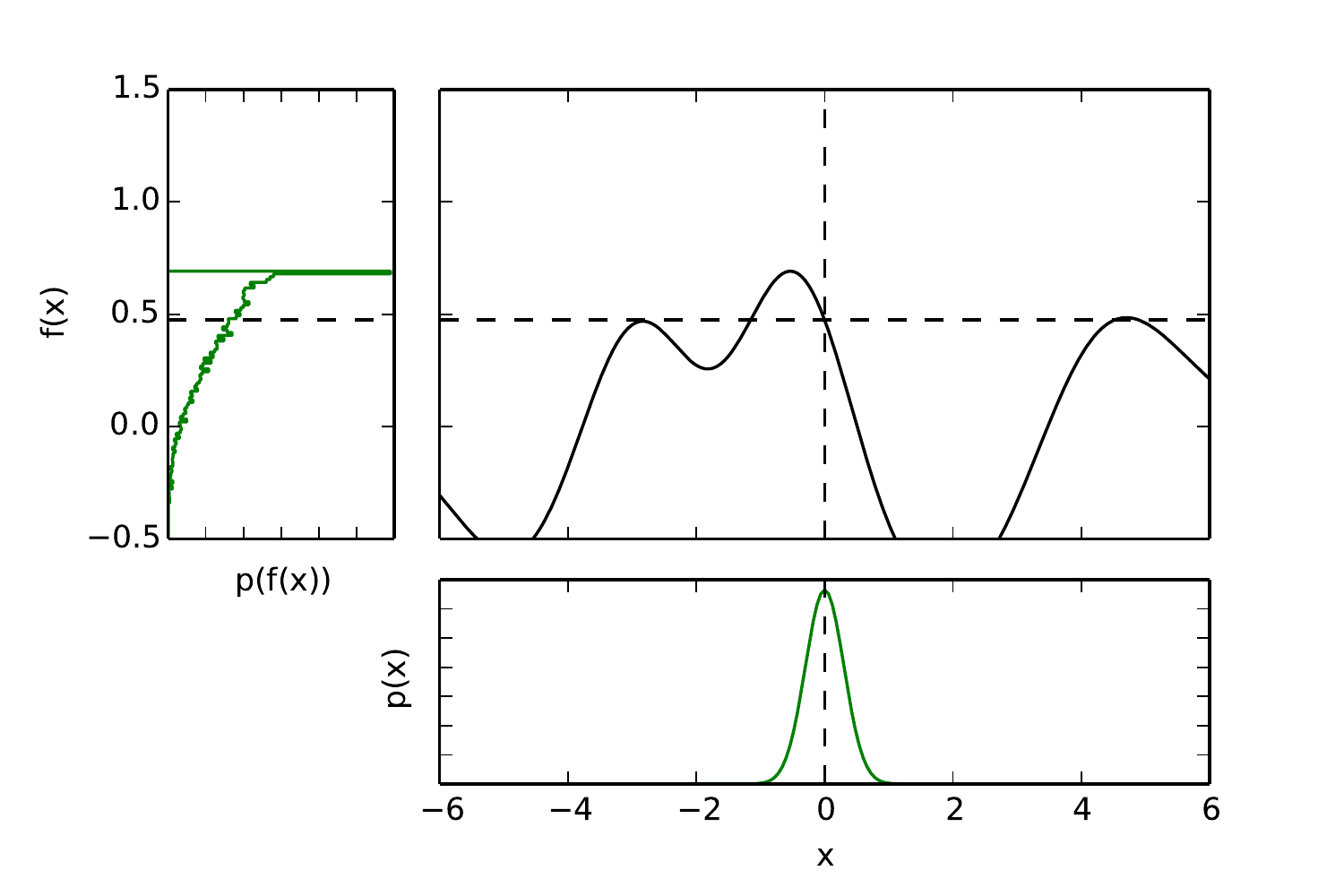}
\caption{Propagation of a Gaussian-distributed input $\slocation$ through a function $f$ sampled from a GP prior. The resulting distribution of output values $f(\slocation)$ is completely different from a Gaussian, since $f$ is highly non-linear. The resulting stochastic process is no longer a Gaussian process.} 
\label{fig:uprop}
\end{figure}

\subsection{Gaussian process priors with uncertain inputs}
\label{sec:ugp}
Consider that, due to measurement and execution noise, when evaluating a function $f$ at a desired input $\location_* \in \R^d$, the actual input location $\tilde{\location}_*|\location_* \sim \pMeasure_*$ where the observation is collected at is not directly observable, but we have access to some estimate of its probability distribution, here indicated by $\pMeasure_*$. Assuming that $f \sim \gp(\mu_0,k)$, the distribution over $\tilde{\fValue}_*=f(\tilde{\location}_*)$ under input noise is no longer Gaussian due to the non-linear relationship between the distribution of $\fValue|\location$, given by Equation \ref{eq:gp_prior},  and $\location$. Figure \ref{fig:uprop} presents an example of what can happen to the output distribution when $f$ is highly non-linear. Therefore, the resulting stochastic process that represents $f$ under random inputs is no longer Gaussian and lacking an analytic formulation \cite{Damianou2016, Girard2004}. However, as demonstrated in \cite{Dallaire2011}, based on the work of \cite{Girard2004}, we can still recover a Gaussian process approximation for $f$ by using the mean and the covariance of the resulting stochastic process under the influence of input noise. In particular, in the case of a constant deterministic mean function $\mu_0(\location) = m_0$, the mean and the covariance of this noisy process are:
\begin{align}
\expectation_{\tilde{\location}_* \sim \pMeasure_*}[f(\tilde{\location}_*)] &= m_0\label{eq:mean_prob}\\
\mathrm{Cov}_{\pMeasure_i, \pMeasure_j}[\tilde{\fValue}_i, \tilde{\fValue}_j]
&= \expectation_{\pMeasure_i,\pMeasure_j}[k(\tilde{\location}_i,\tilde{\location}_j)] ~,
\end{align}
where it is assumed that $\tilde{\location}_i$ and $\tilde{\location}_j$ are independent. Here $\expectation$ denotes \textit{expected value} and $\mathrm{Cov}$ stands for \textit{covariance}. Therefore, the \textit{expected} covariance function is given by:
\begin{equation}
\expectation_{\pMeasure_i,\pMeasure_j}[k(\tilde{\location}_i,\tilde{\location}_j)] = \int_{\tilde{\location}_i \in \R^d} \int_{\tilde{\location}_j \in \R^d} k(\tilde{\location}_i,\tilde{\location}_j) d\pMeasure_i(\tilde{\location}_i)d\pMeasure_j(\tilde{\location}_j)
= k_p(\pMeasure_i,\pMeasure_j) ~.
\label{eq:k_prob}
\end{equation}
Depending on the type of input distributions and the original kernel for deterministic inputs $k(\location,\location')$, approximate and analytical solutions for Equation \ref{eq:k_prob} may exist \cite{Girard2004,Dallaire2011,OCallaghan2012}. An example is the squared exponential kernel in Equation \ref{eq:k_se}. In this case, if the input distributions are Gaussian, the resulting covariance function, according to \cite{Dallaire2011}, is given by:
\begin{equation}
\label{eq:k_pse}
k_p(\normal(\locMean_i,{\Sigma}_i),\normal(\locMean_j,{\Sigma}_j)) = \frac{
\sigma_f^2 \exp\left(-\frac{1}{2} (\locMean_i-\locMean_j)^\transpose (\mathbf{W} + \Sigma_i + {\Sigma}_j)^{-1} (\locMean_i - \locMean_j) \right)
}{
|I + \mathbf{W}^{-1}({\Sigma}_i + {\Sigma}_j)(1-\delta_{ij})|^{1/2}
} ~,
\end{equation}
where $\sigma^2_f$ and $\mathbf{W}$ are the same hyper-parameters as described for the standard squared exponential kernel in Equation \ref{eq:k_se}. 


The posterior of the stochastic process representing $f \sim \gp(\mu_0,k)$ under input noise is not Gaussian due to the complicated forms of Equation \ref{eq:mean} and Equation \ref{eq:var} with respect to $\location$ and $\locSet$. Yet we can still obtain a suitable approximation \cite{Dallaire2011,Girard2004} for the original $f$ in the noisy input setting by doing inference over a GP with mean $m_0$ and covariance function $k_p$, as defined in Equation \ref{eq:k_prob}. This approximation is obtained as:
\begin{align}
\tilde{y}_*|\dataset,\pMeasure_* & \sim \normal(m(\pMeasure_*),v(\pMeasure_*)) \\
\intertext{with}
m(\pMeasure_*) &= m_0 + \mathbf{k}_*^\transpose K_\dataset^{-1}(\observations - \mathbf{m}_0) \label{eq:u_mean}\\
v(\pMeasure_*) &= k_p(\pMeasure_*,\pMeasure_*) - \mathbf{k}_*^\transpose K_\dataset^{-1}\mathbf{k}_* ~, \label{eq:u_var}
\end{align}
where $\mathbf{m}_0$ is an $n$-dimensional mean vector with the constant $m_0$ as elements, $\mathbf{k}_* = [k_p(\pMeasure_*,\pMeasure_1),\dots,k_p(\pMeasure_*,\pMeasure_\nObs)]^\transpose$, $[K_\dataset]_{ij} = k_p(\pMeasure_i,\pMeasure_j) + \delta_{ij} \sigma_n^2$, and $\delta_{ij}$ denotes the Kronecker delta.

\subsection{Bayesian optimisation under localisation uncertainty}
\label{sec:uibo}
To extend BO to the context of uncertain inputs in robotics, we first have to consider how input noise affects the BO process. In robotics problems involving location estimation, we can split input noise into two categories: \textit{localisation} noise and \textit{execution} noise. The first refers to noise affecting the location estimate provided by the robot's localisation system, usually due to imperfections in motion sensing and in other kinds of sensors, such as GPS devices. The second type, execution noise, is the combined effect of everything affecting the execution of the robot's path to a given target location, such as localisation noise, uncertain motion dynamics, etc.

In the BO context, execution noise determines the actual location at which an observation will be taken, while localisation noise affects the estimation of that location. Using an uncertain-inputs GP model, as defined in Section \ref{sec:ugp}, allows us to take into account both execution and localisation noise in the BO algorithm. For instance, assuming additive zero-mean noise, given a target $\location$, the actual location where the observation will be taken is:
\begin{equation}
\tilde{\location}^* = \location + \locNoise_{e} ~,
\end{equation}
where $\locNoise_{e} \sim \pMeasure_e$ represents execution noise. The query location can then be modelled as a random variable $\tilde{\location}^* \sim \pMeasure_\location$. In practice, we usually don't know the execution noise distribution $\pMeasure_e$ exactly, and consequently don't know $\pMeasure_\location$. However, we can use an approximate \textit{querying} distribution $\hat{\pMeasure}_\location \approx \pMeasure_\location$ to account for this uncertainty in the query process. Considering that, at each iteration $t$, the resulting BO loop should select as target location:
\begin{equation}
\location_t = \argmax{\location \in \Sspace} \af(\hat{\pMeasure}_{\location}) ~,
\label{eq:af-p}
\end{equation}
which is computed over the uncertain-inputs GP posterior. Under this formulation, $\hat{\pMeasure}_\location$ provides BO with a measure of how the true location distribution spreads around a given target, due to execution noise. Therefore, any $\hat{\pMeasure}_\location$ that can upper-bound the variance of $\pMeasure_\location$ should be sufficient for the algorithm to work.

After observing the objective function, we assume that BO is provided with another distribution estimating the robot's new location $\tilde{\location}_t \sim \pMeasure_t$, given by localisation, such that its mean estimator is:
\begin{equation}
\hat{\location}_t = \expectation_{\pMeasure_t}[\tilde{\location}_t] = \tilde{\location}_t^* + \locNoise_{l,t} ~,
\end{equation}
where $\locNoise_{l,t} \sim \pMeasure_l$ represents localisation noise. Standard BO would utilise $\hat{\location}_t$ as the input to be added with the corresponding outcome $\observation_t$ into the observations dataset. However, if the localisation noise level is significant or if $\pMeasure_t$ is multi-modal, $\hat{\location}_t$ might not be a good estimator. Fortunately, the GP model in Section \ref{sec:ugp} allows us to use $\pMeasure_t$ directly in the observations dataset, mitigating these effects.

As Equation \ref{eq:af-p} requires, the acquisition function needs to be able to handle probability distributions as inputs by using the uncertain-inputs GP model. For some acquisition functions, such as UCB \cite{Srinivas2012}, it can be as straight forward as replacing $\location$ by $\pMeasure_{\location}$ and using the corresponding GP posterior mean $m(\pMeasure_{\location})$ (Equation \ref{eq:u_mean}) and variance $v(\pMeasure_{\location})$ (Equation \ref{eq:u_var}). For others, some modifications need to be made. In the case of DUCB, a simple but effective modification to the distance-penalty term is to use the distance between the target location and the mean of the distribution of the last sampled location ($\expectation_{\pMeasure_{t-1}}[\tilde{\location}_{t-1}] = \hat{\location}_{t-1}$), i.e.:
\begin{equation}
h_{\text{DUCB}}(\hat{\pMeasure}_\location) = m_{t-1}(\hat{\pMeasure}_\location) + \kappa v^{1/2}_{t-1}(\hat{\pMeasure}_\location) - \gamma d(\hat{\location}_{t-1},\location) ~.
\end{equation}

\section{Experiments}
\label{sec:exp}
In this section, we present experimental results obtained with our method for uncertain-inputs Bayesian optimisation (UIBO) in the presence of localisation uncertainty. In all cases, we used the DUCB acquisition function \cite{Marchant2012} to guide the exploration process. We compare our method against two other approaches. The first is standard BO, which does not consider any uncertainty in the location estimates. The second is unscented Bayesian optimisation (UBO) \cite{Nogueira2016}, which considers execution noise by means of the unscented transform \cite{Wan2000}, but assumes that the location estimate of the observation is accurate.

\subsection{Simulations}
We performed simulations using randomly-generated 2D functions as a model for terrain roughness to be learnt by BO. The task is to find areas of low terrain roughness while staying away from areas of high vibration, which can cause damage to the robot. The learnt GP model should then be more accurate over areas of lower terrain roughness, which are more interesting in practice.

For each test trial, a function is drawn from a Hilbert space \cite{Scholkopf2002} with the reproducing kernel defined as the input-noise-free covariance function and combined with a constant mean to keep vibration values positive. In our case, we used the squared exponential (see Equation \ref{eq:k_se}) and its uncertain-inputs equivalent (Equation \ref{eq:k_pse}) as covariance functions. To simulate input noise, we sample execution noise from a Gaussian $\normal(0,\sigma_e^2 I)$ and localisation noise from another Gaussian $\normal(0,\sigma_l^2 I)$. The querying distribution applied by UIBO is set to also be Gaussian, $\hat{\pMeasure}_\location = \normal(\location,\sigma_\location^2 I)$, and is the same distribution as applied by unscented BO (UBO). We set $\sigma_e = \sigma_l = 0.07$ and $\sigma_\location = 0.1$. Besides input noise, observation noise was set to $\sigma_n = 0.1$. For DUCB's parameters, we set $\kappa = 10$ and $\gamma = 1$ for the three versions of BO, which were manually tuned. 
The hyper-parameters for the GP models were fixed and identical for each BO method.

In the simulations, we also compared each BO approach using maximum entropy search (ES) as heuristics. These heuristics seek only to reduce the entropy of the corresponding GP posterior, choosing to visit areas of high uncertainty to gain information. ES methods should provide a baseline for comparisons, as they do not consider any estimate of the expected vibration in their search. Following ES guidance, the robot should experience high amounts of vibration and long paths. For fair comparisons, the methods based on deterministic-inputs GP models used the equivalent noise-free version of the uncertain-inputs GP covariance function and the same DUCB parameters, since these do not affect the GP model.

\begin{figure}[t]
\centering
\subfloat[Paths and underlying vibration map]{\includegraphics[width=0.9\textwidth]{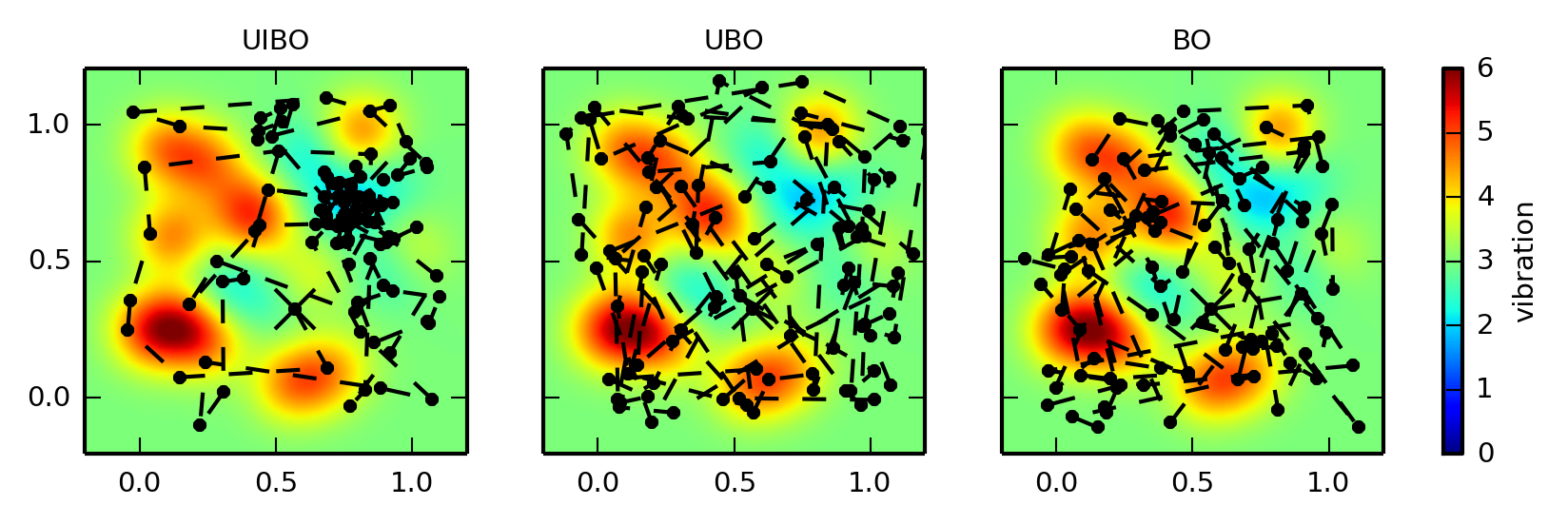}\label{fig:sim_example}} 

\subfloat[Mean experienced vibration ]{\includegraphics[scale=0.8]{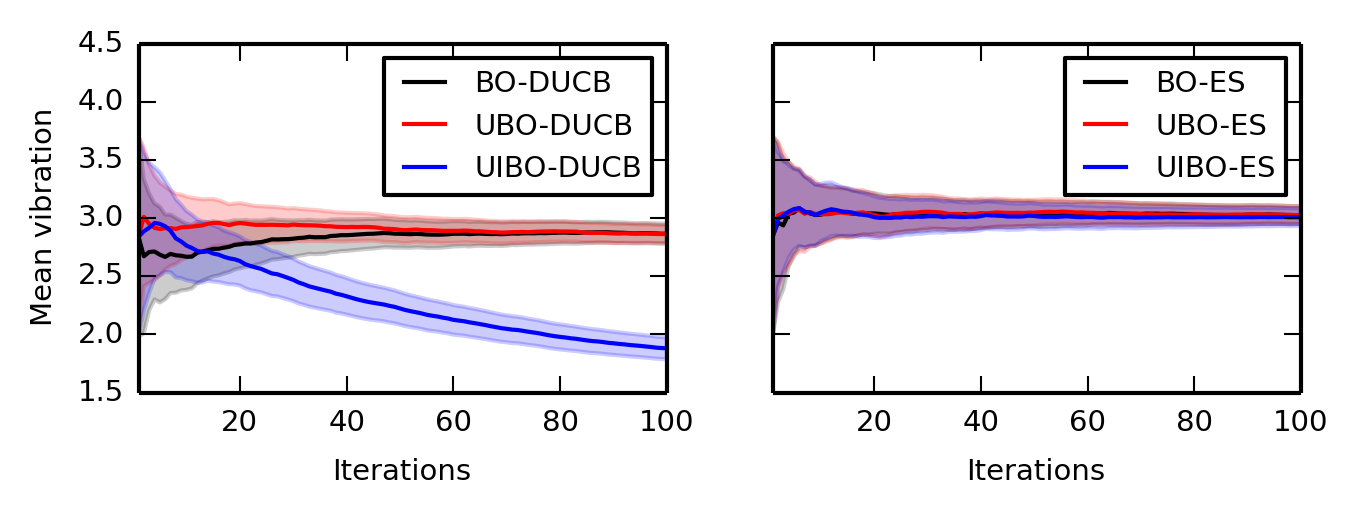}\label{fig:sim_perf}} 

\caption{Simulation results for standard BO, Unscented BO (UBO) and Uncertain-Inputs BO (UIBO) methods for random functions modelling terrain-induced vibration. \protect\subref{fig:sim_example} presents the true paths taken by each BO method using \textit{DUCB} over the true vibration map in one of the test trials. The markers along the paths indicate the locations where observations were taken. The big "X" mark indicates the starting location, which is the same for the three methods at each trial. The plots in \protect\subref{fig:sim_perf} present performance results in terms of mean intensity of the experienced vibration. The results were averaged over 30 trials, each with different maps, and the shaded areas correspond to one standard deviation.}
\label{fig:sim_results}
\end{figure}

Figure \ref{fig:sim_results} presents results obtained in simulation. As seen in the test case in Figure \ref{fig:sim_example}, standard BO and UBO using DUCB end up having a much more exploratory behaviour due to noise, which leads the robot to execute long paths over the terrain. UIBO, on the other hand, is able to focus its exploration on areas of lower vibration intensity. Overall, Figure \ref{fig:sim_perf} shows that UIBO is able to monotonically improve over iterations, finding areas of low vibration for the robot to navigate through. Both the standard and the unscented versions of BO ended up having an average performance much similar to pure entropy search (ES). As entropy-based methods are purely exploratory, there is no improvement in terms of mean experienced vibration, since the algorithm is not optimising for that. In the case of BO-DUCB and UBO-DUCB, the uncertainty in localisation corrupts the location estimates in the observations dataset, and this effect is not taken into account by the GP model. As a result, both methods are misled into areas close to locations that they previously observed, ending up into this more exploratory behaviour, instead of exploiting previous information.

Table \ref{tab:perf} presents a summary of the final performance in terms of different metrics for each method. The root mean square error (RMSE) and weighted RMSE (WRMSE) \cite{Marchant2012} values are computed from the corresponding GP posterior mean ($\mu$) and the ground-truth values of $f$ deterministically queried over a uniform grid covering the entire search space. In our case, the WRMSE emphasises the error in areas of lower vibration, being computed as:
\begin{equation}
WRMSE = \sqrt{\frac{1}{n}\sum_{i=1}^{n} \left((\mu(\location_i) - f(\location_i)) \frac{\max_j f(\location_j) - f(\location_i)}{\max_j f(\location_j) - \min_j f(\location_j)} \right)^2}
\end{equation}
This metric is a better performance indicator for this experiment's task than the plain RMSE, since we are more interested in finding areas of lower vibration.

As seen in Table \ref{tab:perf}, UIBO is able to outperform the other methods when using DUCB, obtaining a good model, in terms of WRMSE, of the terrain roughness while travelling the shortest distance and experiencing the least amount of vibration. Entropy search using the uncertain-inputs GP was able to obtain the smallest RMSE and WRMSE values, but required a much longer distance to be travelled under the cost of high vibration. In general, in terms of relative experienced vibration, BO and UBO methods with DUCB performed worse due to the high amounts of localisation noise, which led them to behave similar to BO-ES. These results indicate that UIBO is able to effectively take into account the uncertainty in localisation, keeping the robot safe while exploring an unknown environment.

\begin{table}[ht]
\centering
\begin{tabular}{|c|c|c|c|c|}
\hline
Method & RMSE & WRMSE & Distance Travelled & Relative Vibration\\
\hline
BO-DUCB & $0.59 \pm 0.09$ & $0.31 \pm 0.05$ & $31.25 \pm 1.98$ & $0.95 \pm 0.04$ \\
\hline
UBO-DUCB & $0.64 \pm 0.12$ & $0.34 \pm 0.08$ & $35.49 \pm 1.23$ & $0.95 \pm 0.04$ \\
\hline
UIBO-DUCB & $0.59 \pm 0.15$ & $0.26 \pm 0.08$ & $\mathbf{16.12 \pm 2.04}$ & $\mathbf{0.61 \pm 0.17}$ \\
\hline
BO-ES & $0.58 \pm 0.10$ & $0.32 \pm 0.07$ & $57.07 \pm 2.73$ & $1.00$ \\
\hline
UBO-ES & $0.64 \pm 0.12$ & $0.34 \pm 0.08$ & $67.15 \pm 2.56$ & $1.00 \pm 0.03$ \\
\hline
UIBO-ES & $\mathbf{0.38 \pm 0.05}$ & $\mathbf{0.22 \pm 0.05}$ & $62.85 \pm 2.06$ & $0.99 \pm 0.03$ \\
\hline
\end{tabular}
\caption{Simulation results: average performance comparisons for each BO approach under different metrics with the corresponding standard deviation.}
\label{tab:perf}
\end{table}


\subsection{Experiments with a real robot}
We performed experiments with a physical robot outdoors to test the performance of the proposed uncertain-inputs BO approach against standard BO with DUCB as in \cite{Souza2014}. The purpose of this experiment is to highlight the differences in performance and in behaviour between both BO approaches when faced with localisation uncertainty in a real-world scenario, where the robot is tasked with learning terrain roughness.

\subsubsection{Experimental setup}
\begin{figure}[t] 
\centering
\subfloat[Robot platform]{\includegraphics[width=0.36\textwidth]{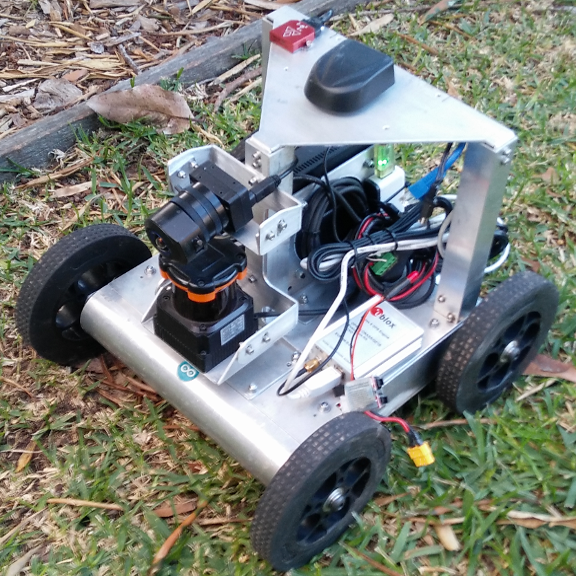}\label{fig:robot}}
\subfloat[Vibration measurements]{\includegraphics[width=0.55\textwidth]{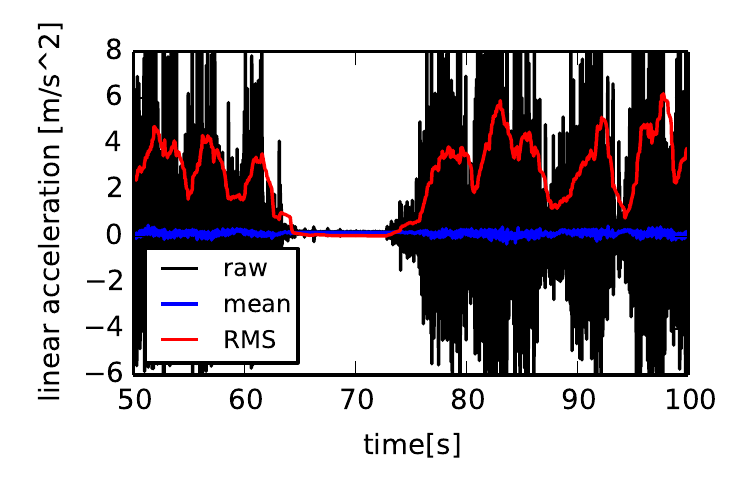}\label{fig:rms}}
\caption{Robot experiment details. \protect\subref{fig:robot} presents our experimental platform. \protect\subref{fig:rms} compares different methods to measure vibration using an IMU: \textit{raw} corresponds to the raw vertical acceleration readings (discounting gravity), \textit{mean} corresponds to the mean value of the raw measurements in a moving window of size 200, and \textit{RMS} corresponds to the root-mean-square value of the raw measurements in the same window.}
\end{figure}

Our test platform was a small four-wheeled skid-steer robot, depicted in Figure \ref{fig:robot}. The robot is equipped with an on-board computer running ROS \footnote{The Robot Operating System: \url{www.ros.org}}. The tests were performed in an area with terrain covered by grass and with some portions of harder ground exposed. The goal of the robot is to find areas of low terrain roughness, while avoiding areas that induce excessive vibration to the platform. In this scenario, high amounts of vibration can cause damage to the robot or affect its ability to perceive the environment. 


Our robot was equipped with an IMU sensor, placed on top of its chassis. With the robot moving, vibration shows up mostly as linear acceleration along the robot's vertical axis, bouncing the robot up and down to the ground. We considered different methods to measure vibration from a fixed moving window of acceleration measurements. In particular, we considered the mean and the root mean square values. As we want the robot to avoid areas of high vibration, the RMS value demonstrated itself to be a better estimator, since it grows with the amplitude of the vibration, as pictured in Figure \ref{fig:rms}. The mean, on the other hand, should be always around zero, as each upwards acceleration is immediately compensated by a downwards drop of the robot.


As observations for the BO algorithms, the RMS vibration estimates were computed over a sliding window of 100 IMU measurements and updated at a rate of 2 Hz as the robot is driving. To reduce the effect of different driving speeds on the readings, the vibration estimates were posted when the robot was driving at speeds between 0.4 and 0.7 m/s, where 0.7 m/s was the maximum speed allowed for the path following control. These observations were combined with location estimates from an extended Kalman filter (EKF) \cite{MooreStouchKeneralizedEkf2014}, which was configured to fuse wheel odometry, IMU and GPS estimates.   

We ran both plain BO and UIBO for a fixed budget of 30 iterations, i.e. each algorithm was allowed to choose 30 target locations to take the robot to. At each iteration, the robot drove autonomously attempting to follow straight paths from its previous goal to the next goal, as given by the BO planner. The iteration was signalled as finished whenever the robot arrived within a given radius from the goal. After that, the GP model is updated, the hyper-parameters are re-learnt by maximising the GP log-marginal likelihood using COBYLA \cite{Powell2007}, and the planner selects a new target. As the sample rate of observations is relatively low, both BO methods were able to repeatedly execute this process online in close to real time. The DUCB acquisition function was configured with an uncertainty factor $\kappa=10$ and a distance factor $\gamma=0.5$. UIBO applied as query distribution an isotropic Gaussian with variance $\sigma^2_\location=4$ for each coordinate.

\subsubsection{Performance}
\begin{figure}[t]
\centering
\includegraphics[height=4cm]{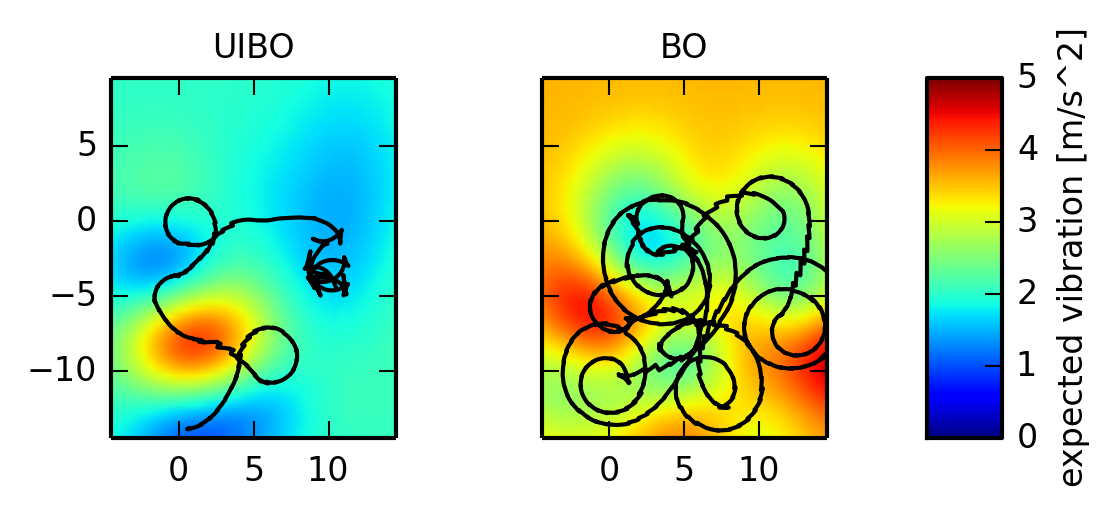}
\caption{Posterior mean of the GP model built by each BO method overlaid with their respective paths according to locations estimated by an EKF fusing conventional GPS, IMU and odometry.}
\label{fig:field-gp}
\end{figure}

Figure \ref{fig:field-gp} presents the GP model obtained by each BO method and their paths, according to noisy EKF location estimates. For both methods, the robot was placed at an initial location at the bottom left corner of the search space. As Figure \ref{fig:field-gp} shows, UIBO is able to concentrate its search over areas of low vibration. On the other hand, standard BO ends up performing too much exploration passing more often over areas that cause excessive vibration to the robot.

\begin{figure}[t]
\centering
\includegraphics[height=4cm]{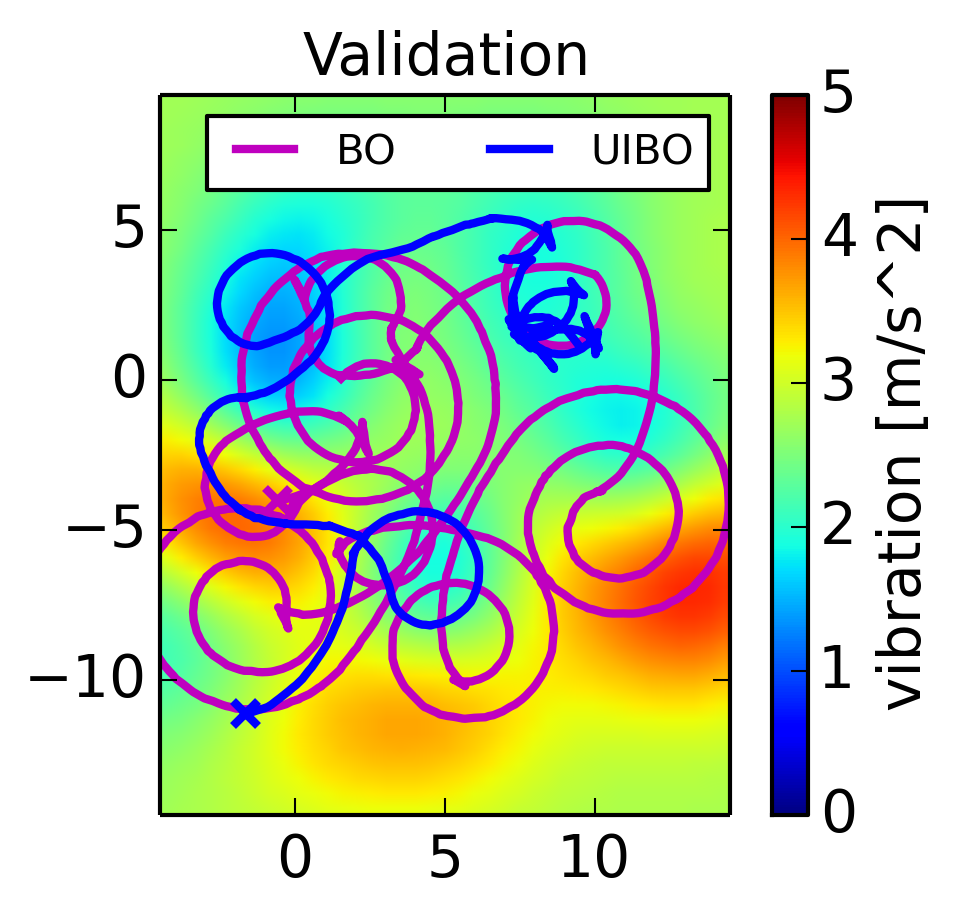}
\caption{For visualisation purposes, posterior mean of the GP model built from the validation data overlaid with the true paths taken by each method, according to RTK GPS location estimates.}
\label{fig:field-validation}
\end{figure}

For performance comparisons, we collected a set of vibration measurements during the experiments using a real time kinematic (RTK) GPS \footnote{\url{https://emlid.com/reach/}} device, which was employed to obtain high-precision location estimates. These measurements were only collected for the purposes of validation, and not passed on to the BO planners, which relied solely on a conventional GPS device fused into the EKF estimates. The uncertainty of the EKF location estimates varied around 3 metres, while the RTK GPS device is able to provide centimetre accuracy. The estimates of a GP model built directly from the validation data is shown in Figure \ref{fig:field-validation}. This model is presented for visualisation purposes and was not used to compute performance metrics, which relied only on the raw validation data. The model's hyper-parameters were determined by maximising its log-marginal likelihood as $\{m_0,l,\sigma_f,\sigma_n\} = \{2.7,2.8,0.8,1.0\}$, where $l$ is the kernel length-scale.

Table \ref{tab:field-perf} presents a performance summary. The WRMSE value for this experiment was computed between the posterior mean from each method's final GP model, built from noisy EKF location estimates, and the raw vibration measurements present in the validation dataset, collected using precise RTK GPS location estimates. We can see that, for this experiment, UIBO is able to outperform standard BO by producing a better model over areas of low roughness while experiencing less vibration and following a much shorter path, confirming facts previously observed in simulations. 



\begin{table}[t]
\centering
\begin{tabular}{|c|c|c|c|c|}
\hline
Method & WRMSE [$m/s^2$] & RMSE [$m/s^2$] & Distance Travelled [$m$] & Mean Vibration [$m/s^2$]\\
\hline
BO & 1.04 & 1.31 & 247 &  3.0
 \\
\hline
UIBO & 0.83 & 1.33 & 92 & 2.1 \\
\hline
\end{tabular}
\caption{Field results: performance comparisons for each BO approach under different metrics. The WRMSE was computed between the posterior mean of the final GP model at the locations in the validation dataset and the corresponding vibration measurements at those locations.}
\label{tab:field-perf}
\end{table}

\section{Conclusion}
\label{sec:con}
In this paper, we presented a new method to Bayesian optimisation for active learning problems in robotics in the cases where uncertainty in the location estimates is significant. The proposed method provides a principled way to consider this uncertainty in the inputs of a Gaussian process model, which is applied as a prior by the BO algorithm. Execution noise is also considered by means of a query distribution when optimising BO's acquisition function. The method was proved to outperform other BO approaches in simulation and in an experiment with a physical robot. Therefore, the proposed method can be applied to problems where we are interested in mapping the traversability of a terrain, but need to keep the robot safe in the midst of localisation uncertainty. 

Some topics were not addressed by this paper but are worthy of future research work. One of them is the estimation of execution noise, which could be done in an online way by other statistical methods, for example, using maximum likelihood estimates for the parameters of a given distribution. Another topic is the extension of the DUCB-based exploration to informative continuous path planning \cite{Marchant2014} or to trajectory optimisation in goal-directed navigation \cite{Oliveira2016}. For that, however, a method to propagate execution noise through a candidate path would have to be developed taking into account hard-to-model factors from both the robot and the environment, such as stochastic motion dynamics and imperfections in sensor noise.

\acknowledgement{This work was partly funded by CAPES, Brazil, scholarship BEX 13224/13-1, and by Data61/CSIRO, Australia.}

\bibliographystyle{splncs03}
\bibliography{references}

\end{document}